\documentclass{article}
\usepackage[final,nonatbib]{neurips_2022}

\usepackage[utf8]{inputenc} 
\usepackage[T1]{fontenc}    
\usepackage{xcolor}         

\usepackage[breaklinks=true,colorlinks,bookmarks=true]{hyperref}
\usepackage{url}            
\usepackage{booktabs}       
\usepackage{amsfonts}       
\usepackage{nicefrac}       
\usepackage{microtype}      
\usepackage{algorithm}
\usepackage{algorithmic}
\usepackage{multirow}
\usepackage{bm}
\usepackage{amsmath,amssymb,amsfonts}
\usepackage{epsfig}
\usepackage{colortbl}
\usepackage{wrapfig}

\usepackage{amsthm}

\usepackage{tablefootnote}

\usepackage{pifont}
\usepackage{marvosym}

\begin{document}

\title{Towards Theoretically Inspired Neural Initialization Optimization}







\author{Yibo Yang$^{1}$, Hong Wang$^{2}$, Haobo Yuan$^3$, Zhouchen Lin$^{2,4,5*}$ \vspace{2mm}\\
	\small{$^1$JD Explore Academy, Beijing, China}\\
	\small{$^2$Key Lab. of Machine Perception (MoE), School of Intelligence Science and Technology, Peking University}\\
	\small{$^3$Institute of Artificial Intelligence and School of Computer Science, Wuhan University}\\
	\small{$^4$Institute for Artificial Intelligence, Peking University}\\
	\small{$^5$Pazhou Laboratory, Guangzhou, China}
}


\maketitle

{\let\thefootnote\relax\footnotetext{$*$: corresponding author.}}

\vspace{-4mm}
\begin{abstract}
	\vspace{-1mm}
Automated machine learning has been widely explored to reduce human efforts in designing neural architectures and looking for proper hyperparameters. In the domain of neural initialization, however, similar automated techniques have rarely been studied. Most existing initialization methods are handcrafted and highly dependent on specific architectures. In this paper, we propose a differentiable quantity, named GradCosine, with theoretical insights to evaluate the initial state of a neural network. Specifically, GradCosine is the cosine similarity of sample-wise gradients with respect to the initialized parameters. By analyzing the sample-wise optimization landscape, we show that both the training and test performance of a network can be improved by maximizing GradCosine under gradient norm constraint. Based on this observation, we further propose the neural initialization optimization (NIO) algorithm. Generalized from the sample-wise analysis into the real batch setting, NIO is able to automatically look for a better initialization with negligible cost compared with the training time. With NIO, we improve the classification performance of a variety of neural architectures on CIFAR-10, CIFAR-100, and ImageNet. Moreover, we find that our method can even help to train large vision Transformer architecture without warmup. 
\end{abstract}





%
\vspace{-2mm}
\section{Introduction}
\label{intro}
\vspace{-1mm}

For a deep neural network, architecture \cite{he2016deep,howard2017mobilenets,huang2017densely} and parameter initialization \cite{he2015delving,glorot2010understanding} are two initial elements that largely account for the final model performance. Lots of human efforts have been devoted to finding better answers with respect to the two aspects. To automatically produce better architectures, neural architecture search \cite{zoph2016neural,real2019regularized,chen2018searching} has been a research focus. However, on the other hand, using automated techniques for parameter initialization has rarely been studied. 


Previous initialization methods are mostly handcrafted. They focus on finding proper variance patterns of randomly initialized weights \cite{he2015delving,glorot2010understanding,saxe2013exact,mishkin2015all} or rely on empirical evidence derived from certain architectures \cite{zhang2019fixup, huang2020improving, gilmer2021loss,das2021data}. Recently, \cite{zhu2021gradinit,dauphin2019metainit} propose learning-based initialization that learns to tune the norms of the initial weights so as to minimize a quantity that is intimately related to favorable training dynamics. Despite being architecture-agnostic and gradient-based,
these methods do not consider sample-wise landscape.
Their derived quantities lack theoretical supports for being related to model performance. It is unclear whether optimizing these quantities can indeed lead to better training or generalization performance. 


In order to find a better initialization, a theoretically sound quantity that intimately evaluates both the training and test performance should be designed. Finding such a quantity is a non-trivial problem and shows many challenges. First, the test performance is mostly decided by the converged parameters after training, while the training dynamic is more related to the parameters at initialization or during training. 
Second, to efficiently find a better starting point, the quantity is expected to be differentiable to enable the optimization in the continuous parameter space. 
To this end, we leverage the recent advances in optimization landscape analysis \cite{allen2019convergence} to propose a novel differentiable quantity and develop a corresponding algorithm for its optimization at initialization. 




Specifically, our quantity is inspired by analyzing the optimization landscapes of individual training samples \cite{zhang2021gradsign}. Through generalizing prior theoretical results on batch-wise optimization \cite{allen2019convergence} to sample-wise optimization, we prove that both the network’s training and generalization error are upper bounded by a theoretical quantity that correlates with the cosine similarity of the sample-wise local optima. Moreover, this quantity also relates to the training dynamic since it reflects the optimization path consistency \cite{meuleau2002ant,oymak2019overparameterized} from the starting point. Unfortunately, the sample-wise local optima are intractable. With the hypothesis that the sample-wise local optima can be reached by the first-order approximation from the initial parameters, we can approximate the quantity via the sample-wise gradients at initialization.
Our final result shows that, under a limited gradient norm, both the training and test performance of a network can be improved by maximizing the cosine similarity of sample-wise gradients, named GradCosine, which is differentiable and easy to implement. 

We then propose the Neural Initialization Optimization (NIO) algorithm based on GradCosine to find a better initialization agnostic of architecture. We generalize the algorithm from the sample-wise analysis into the batch-wise setting by dividing a batch into sub-batches for friendly implementation. We follow \cite{dauphin2019metainit,zhu2021gradinit} using gradient descent to learn a set of scalar coefficients of the initialized parameters. These coefficients are optimized to maximize GradCosine for better training dynamic and expected performance while constraining the gradient norm from explosion. 


Experiments show that for a variety of deep architectures including ResNet \cite{he2016deep}, DenseNet \cite{huang2017densely}, and WideResNet \cite{zagoruyko2016wide}, our method achieves better classification results on CIFAR-10/100 \cite{krizhevsky2009learning} than prior heuristic \cite{he2015delving} and learning-based \cite{dauphin2019metainit,zhu2021gradinit} initialization methods. We can also initialize ResNet-50 \cite{he2016deep} on ImageNet \cite{deng2009imagenet} for better performance. 
Moreover, our method is able to help the recently proposed Swin-Transformer \cite{liu2021swin} achieve stable training and competitive results on ImageNet even without warmup \cite{goyal2017accurate}, which is crucial for the successful training of Transformer architectures \cite{liu2019variance,xiong2020layer}. 

\section{Related Work}
\subsection{Network Initialization}
Existing initialization methods are designed to control the norms of network parameters via Gaussian initialization \cite{glorot2010understanding,he2015delving} or orthonormal matrix initialization \cite{saxe2013exact,mishkin2015all} with different variance patterns. These analyses are most effective for simple feed-forward networks without skip connections or normalization layers. Recently, initialization techniques specified for some complex architectures are proposed. For example, \cite{zhang2019fixup} studied how to initialize networks with skip connections and \cite{huang2020improving} generalized the results into Transformer architecture \cite{vaswani2017attention}. However, these heuristic methods are restricted to specific architectures. Automated machine learning has achieved success in looking for hyperparameters \cite{bergstra2011algorithms,feurer2019hyperparameter} and architectures \cite{zoph2016neural,real2019regularized,chen2018searching,yang2020ista,yang2021towards,huang2020explicitly}, while similar techniques for neural network initialization deserve more exploration. Current learning based initialization methods \cite{dauphin2019metainit,zhu2021gradinit} optimize the curvature \cite{dauphin2019metainit} or the loss reduction of the first stochastic step \cite{zhu2021gradinit} using gradient descent to tune the norms of the initial parameters. 
However, these methods lack theoretical foundations to be related to the model performance. Different from these methods, our proposed GradCosine is derived from a theoretical quantity that is the upper bound of both training and generalization error.

\subsection{Evaluating Model Performance at Initialization}
Evaluating the performance of a network at initialization has been an important challenge and widely applied in zero-shot neural architecture search \cite{abdelfattah2020zero,mellor2021neural,chen2021neural,simon2021neural,zhang2021gradsign,shu2021nasi,li2018optimization} and pruning \cite{wang2020picking,lee2018snip, tanaka2020pruning}. The evaluation quantities in these studies are mainly based on the initial gradient norm \cite{tanaka2020pruning,shu2021nasi}, the eigenvalues of neural tangent kernel \cite{chen2021neural,shu2021nasi}, and the Fisher information matrix \cite{turner2019blockswap,turner2021neural,theis2018faster}. However, these quantities cannot reflect optimization landscape, which is crucial for training dynamic and generalization \cite{brutzkus2017globally,du2018gradient,ge2017learning,soltanolkotabi2017learning,li2017convergence,allen2019convergence}. \cite{allen2019convergence} provided theoretical evidence that for a sufficiently large neighborhood of a random initialization, the optimization landscape is nearly convex and semi-smooth. Based on the result, \cite{zhang2021gradsign} proposed to use the density of sample-wise local optima to evaluate and rank neural architectures. Our study also performs sample-wise landscape analysis, but differs from \cite{zhang2021gradsign} in that our proposed quantity is differentiable and reflects optimization path consistency, while the quantity in \cite{zhang2021gradsign} is non-differentiable so cannot serve for initialization optimization. We make a comparison in more details in Appendix \ref{explanation}.

\begin{figure}[t]
	\vskip 0.2in
	\begin{center}
		\centerline{\includegraphics[width=\columnwidth]{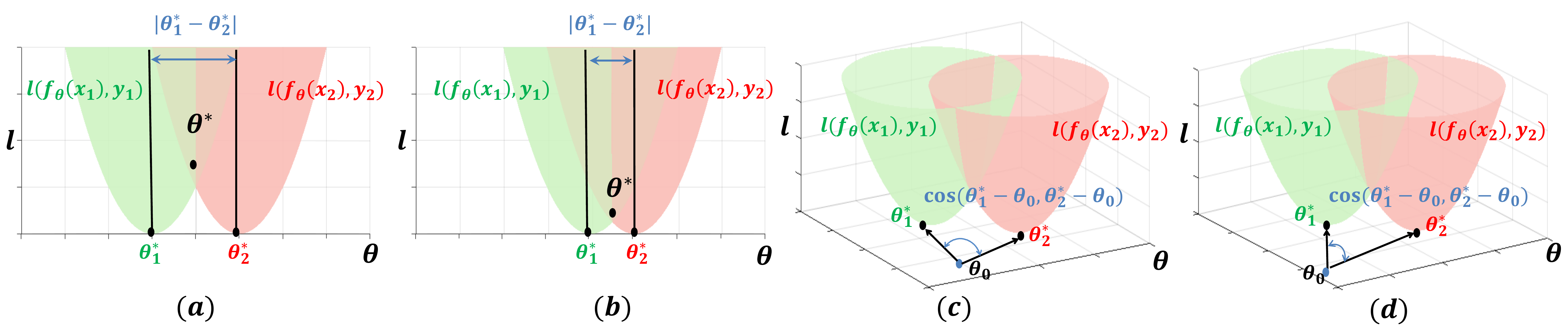}}
		\caption{(a) Optimization landscape with sparser sample-wise local optima corresponds to a worse $\theta^*$ (larger loss $l$). (b) Optimization landscape with denser sample-wise local optima corresponds to a better $\theta^* $ (smaller loss $l$). However, the density of sample-wise local optima cannot reflect training path. We further leverage the cosine similarity of sample-wise local optima.  Under the same local optima density, (d) corresponds to a better training dynamic than (c), since (d) enjoys a better optimization path consistency (smaller cosine distance between $\theta^*_1-\theta_0$ and $\theta^*_2-\theta_0$). We give more detailed explanations and discussions in Appendix \ref{explanation}.}
		\label{fig1}
	\end{center}
	\vspace{-3mm}
\end{figure}

\section{Theoretical Foundations}
\label{theory}
\subsection{Sample-wise Optimization Landscape Analysis}
Conventional optimization landscape analyses \cite{li2017convergence,allen2019convergence} mainly focus on the objective across a mini-batch of samples and miss potential evidence hidden in the optimization landscapes of individual samples. As recently pointed out in \cite{zhang2021gradsign}, by decomposing a mini-batch objective into the summation of sample-wise objectives across individual samples in the mini-batch, they find that the network with denser sample-wise local optima tends to reach a better local optima for the mini-batch objective, as shown in the Figure (\ref{fig1}) (a)-(b). Based on this insight, they propose to use sample-wise local optima density to judge model performance at initialization. 



\textbf{Density of sample-wise local optima.}
For a batch of training samples $S = \{(x_i,y_i)\}_{i\in[n]}$, a loss function $l(\hat y_i, y_i)$, and a network $f_\theta(\cdot)$ parameterized by $\theta\in\mathbb{R}^m$, the sample-wise local optima density $\Psi_{S,l}(f_{\theta_0}(\cdot))$ is measured by the averaged Manhattan distance between the pair-wise local optima $\{\theta^*_i\}_{i\in[n]}$ of all $n$ samples near the random initialization $\theta_0$ \cite{zhang2021gradsign}, \emph{i.e.,}
\begin{equation}\label{gradsign}
    \Psi_{S,l}(f_{\theta_0}(\cdot)) = \frac{\sqrt{\mathcal{H}}}{n^2}\sum_{i,j}||\theta^*_i-\theta^*_j||_1,\ \ i,j\in[1,n],
\end{equation}
where $\mathcal{H}$ is the smoothness upper bound: $\forall k\in [m], i\in [n], [\nabla^2l(f_{\theta}(x_i), y_i)]_{k,k}\leq\mathcal{H}$ and $\mathcal{H}$ always exists with the smoothness assumption that for a neighborhood $\Gamma_{\theta_0}$ of a random initialization $\theta_0$, the sample-wise optimization landscapes are nearly convex and semi-smooth \cite{allen2019convergence}. Based on this assumption, the training error of the network can be upper bounded by $\frac{n^3}{2}\Psi^2_{S,l}(f_{\theta_0}(\cdot))$. Moreover, with probability $1-\delta$, the generalization error measured by population loss $\mathbb{E}_{(x_u,y_u)\sim\mathcal{D}}[l(f^*_\theta(x_u), y_u)]$ is upper bounded by $\frac{n^3}{2}\Psi^2_{S,l}(f_{\theta_0}(\cdot))+\frac{\sigma}{\sqrt{n\delta}}$, where $\mathcal D=\{(x_u,y_u)\}_{u\in[U]}$ is the underlying data distribution of test samples and $\sigma^2$ is the upper bound of $Var_{(x_u,y_u)\sim\mathcal D} [||\theta^*-\theta^*_u||^2_1]$ \cite{zhang2021gradsign}. 



Although the sample-wise local optima density is theoretically related to both the network training and generalization error, we point out that it may not be a suitable quantity to evaluate network initialization due to the following reasons.
First, it ignores the optimization path consistency from initialization $\theta_0$ to each sample-wise local optimum $\theta^*_i$. As shown in Figure \ref{fig1} (c)-(d), while both (c) and (d) have the same local optima density, in Figure \ref{fig1} (d) the optimization path from the initialization to the local optima of the two samples are more consistent. Training networks on samples with more consistent optimization paths using batch gradient descent naturally corresponds to a better training dynamic that enjoys faster and more stable convergence  \cite{ruder2016overview,oymak2019overparameterized}. Second, the sample-wise local optima in Eq. (\ref{gradsign}) are intractable. It can be approximated by measuring the consistency of sample-wise gradient signs \cite{zhang2021gradsign}. But it is non-differentiable and cannot serve for initialization optimization. 

Based on this observation, we aim to derive a new quantity that directly reflects optimization path consistency and is a differentiable function of the initialization $\theta_0$.
Our proposed quantity is based on the cosine similarity of the paths from initialization to sample-wise local optima.

\textbf{Cosine similarity of sample-wise local optima.}
Concretely, our quantity can be formulated as:
\begin{equation}
    \Theta_{S,l}(f_{\theta_0}(\cdot)) = \frac{\mathcal{H}\alpha^2}{n}\sum_{i,j}\left(\frac{\alpha}{\beta} - \cos\angle(\theta^*_i-\theta_0,\theta^*_j-\theta_0)\right),\ \ i,j\in[1,n],
    \label{eq2}
\end{equation}
where $\alpha$ and $\beta$ are the maximal and minimal $\ell_2$-norms of the sample-wise optimization paths, \emph{i.e.,} $\alpha=\max(||\theta^*_i-\theta_0||_2), \beta=\min(||\theta^*_i-\theta_0||_2), \forall i \in [1,n]$,
and $\cos\angle(\theta^*_i-\theta_0,\theta^*_j-\theta_0)$ refers to the cosine similarity of the paths from initialization $\theta_0$ to sample-wise local optima, $\theta^*_i$ and $\theta^*_j$. The cosine term in Eq. (\ref{eq2}) reflects the optimization path consistency. Together with the distance term $\frac{\alpha}{\beta}$, it is also able to measure the density of sample-wise local optima. 
In the ideal case, when all the local optima are located at the same point, $\Theta_{S,l}(f_{\theta_0}(\cdot))=\Psi_{S,l}(f_{\theta_0}(\cdot))=0$. Hence, compared with $\Psi$, our $\Theta$ is more suitable for evaluating the initialization quality.

\subsection{Main Results}
\label{s32}
In this subsection, we theoretically illustrate how minimizing the quantity $\Theta$ in Eq. (\ref{eq2}) corresponds to better training and generalization performance. Similar to \cite{zhang2021gradsign}, our derivations are also based on the evidence that there exists a neighborhood for a random initialization such that the sample-wise optimization landscapes are nearly convex and semi-smooth \cite{allen2019convergence}.

\textbf{Lemma 1.} \textit{There exists no saddle point in a sample-wise optimization landscape and every local optimum is a global optimum \cite{zhang2021gradsign}.}

Based on Lemma 1, we can draw a relation between the training error and $\Theta_{S,l}(f_{\theta_0}(\cdot))$. Moreover, we show that the proposed quantity is also related to  generalization performance as the upper bound of population error. We present the following two theoretical results.

\textbf{Theorem 2.} \textit{The training loss $\mathcal{L}=\frac{1}{n}\sum_i l(f_{\theta^*}(x_i), y_i)$ of a trained network $f_{\theta^*}$ on a dataset $S = \{(x_i,y_i)\}_{i\in[n]}$ is upper bounded by $\Theta_{S,l}(f_{\theta_0}(\cdot))$, and the bound is tight when $\Theta_{S,l}(f_{\theta_0}(\cdot))=0$.}

\textbf{Theorem 3.} \textit{Suppose that $\sigma^2$ is the upper bound of $Var_{(x_u,y_u)\sim\mathcal D} [||\theta^*-\theta^*_u||^2_2]$, where $\theta^*_u$ is the local optimum in the convex neighborhood of $\theta_0$ for test sample $(x_u, y_u)$. With probability $1-\delta$, the population loss $\mathbb{E}_{(x_u,y_u)\sim\mathcal{D}}[l(f_{\theta^*}(x_u), y_u)]$ is upper bounded by $\Theta_{S,l}(f_{\theta_0}(\cdot))+\frac{\sigma}{\sqrt{n\delta}}$}.

We provide proofs of these two theorems in Appendix \ref{ap1}. Combining both theorems, we can conclude that $\Theta_{S,l}(f_{\theta_0}(\cdot))$ is the upper bound of both training and generalization errors of the network $f_{\theta^*}$. Therefore, minimizing $\Theta$ theoretically helps to improve the model performance.






Albeit theoretically sound, $\Theta_{S,l}(f_{\theta_0}(\cdot))$ requires sample-wise optima $\theta^*_i$ which are intractable at initialization. To this end, we will show how to develop a differentiable and tractable objective based on Eq. (\ref{eq2}) in Section \ref{sec_GC}, and introduce the initialization optimization algorithm in Section \ref{sec_NIO}. 


\section{GradCosine}
\label{sec_GC}

\subsection{First-Order Approximation of Sample-Wise Optimization}
\label{s42}
Since we are dealing with sample-wise optimization, it is reasonable to calculate each sample-wise optimum by the first-order approximation. We hypothesize that each sample-wise optimum can be reached via only one-step gradient descent from the initialized parameters. Its rationale lies in that it is very easy for a deep neural network to learn the optimum for only one training sample with gradient descent. Based on this hypothesis, we can approximate each local optimum as:
\begin{equation}
    \theta^*_i \approx \theta_0 - \eta g_i,\ \ i\in[1,n],
    \label{eq4}
\end{equation}
where $g_i=\nabla_\theta l(f_\theta(x_i), y_i)|_{\theta_0}$ is the sample-wise gradient at initialization, and $\eta$ is the learning rate. With Eq. (\ref{eq4}), the upper bound quantity in Eq. (\ref{eq2}) can be simplified as:
\begin{equation}
	\label{eq5}
		\Theta \approx
		\frac{\mathcal{H}g_{max}^2}{n}\sum_{i,j}\left(\frac{g_{max}}{g_{min}} - \cos\angle(g_i,g_j)\right),\ \ i,j\in[1,n],
\end{equation}

where $g_{max}$ and $g_{min}$ are the maximal and minimal sample-wise gradient norms at initialization, \emph{i.e.,} $g_{max} = \max(||g_i||_2), g_{min} = \min(||g_i||_2), \forall i\in[1,n]$.


\subsection{Guidance for Initialization}

Suppose that the sample-wise gradient at initialization is upper bounded by a constant $\gamma$, \emph{i.e.,} $g_{max}\le \gamma$, and then we have:
\begin{equation}
	\label{relax_theta}
	\Theta \le
	\frac{\mathcal{H}\gamma^2}{n}\sum_{i,j}\left(\frac{\gamma}{g_{min}} - \cos\angle(g_i,g_j)\right),\ \ i,j\in[1,n].
\end{equation}
From Eq. (\ref{eq2}) to Eq. (\ref{relax_theta}), we have almost converted the upper bound into a quantity that can be easily calculated by the sample-wise gradient at initialization, except for the term $g_{min}$ and the constraint $g_{max}\le \gamma$. We will explicitly add corresponding optimization objective and constraint in our neural initialization optimization algorithm in Section \ref{sec_NIO} to ensure a small $\gamma/g_{min}$.



From Eq. (\ref{relax_theta}), we can draw some useful guidance for a better initialization:
\begin{itemize}
    \item[(1)] The sample-wise gradient norms of the initialized parameters should be large and close to the maximal value bounded by $\gamma$ to induce a small $\gamma/g_{min}$;
    \item[(2)] The cosine similarity of the sample-wise gradients should be as large as possible.
\end{itemize}

\begin{algorithm}[t]
		\small
\caption{
GradCosine (GC) and gradient norm (GN)
}
\label{alg1}
\begin{algorithmic}[1]
\REQUIRE Initial network parameters $\theta_0$, and a batch of samples $S=\{(x_i,y_i)\}_{i\in[B]}, B = |S|$.
\FOR{i=1 to $B$}
\STATE Compute the sample-wise gradient:\\ $g_i \leftarrow \nabla_\theta l(f_\theta(x_i), y_i)|_{\theta_0}; g_i \in \mathbb{R}^m$
\ENDFOR
\STATE Compute the average of gradient norm:\\ $\textbf{GN}(S,\theta_0)\leftarrow \frac{1}{B}\sum^B_{i=1}||g_i||_2$
\STATE Compute the cosine similarity of gradients:\\ $\phi_{i,j} \leftarrow \frac{g_i\cdot g_j}{||g_i||_2\cdot||g_j||_2}, i=1,...,B, j=1,...,B$
\STATE Compute the average of gradient cosine: \\
$\textbf{GC}(S,\theta_0)\leftarrow \frac{1}{B^2}\sum^{B}_{i=1}\sum^{B}_{j=1}\phi_{i,j}$
\STATE Output \textbf{GN} and \textbf{GC}.
\end{algorithmic}
\end{algorithm}

\textbf{Relations to Favorable Training Dynamic.}
There are significant evidences in existing studies that support the two rules for better training dynamic. The first rule is intimately related to the neural tangent kernel (NTK) \cite{jacot2018neural} that has recently been shown to determine the early learning behavior \cite{xiao2020disentangling,shu2021nasi,chen2021neural,simon2021neural}. Specifically, \cite{shu2021nasi} finds that the training dynamic of a neural network can be characterized by the trace norm of NTK at initialization and further approximates it via the initial gradient norm as a gauge to search for neural architectures. \cite{zhu2021gradinit} initializes a network by maximizing the loss reduction of the first gradient descent step, which also corresponds to gradient norm in the first-order approximation of the loss function. The two rules together prefer a model whose sample-wise gradients have close $\ell_2$ and cosine distances. It is in line with prior observations that the initial gradient variance should be small to enable a large learning rate \cite{liu2019variance,zhuang2020adabelief}. However, gradient variance is not a proper optimization objective due to its high sensitivity to gradient norm. 
 




While the first rule of improving the initial gradient norm has been suggested for better training dynamic in prior studies \cite{zhu2021gradinit,shu2021nasi},
the second rule is completely new. We refer this novel quantity, \emph{i.e.,} $\cos\angle(g_i, g_j)$, for evaluating the network initialization as GradCosine. Intuitively, in the ideal case where the initial gradients of all samples are identical, we have the smallest upper bound of training and generalization error. Moreover, both gradient norm and GradCosine are differentiable and thus enable optimization to find a better initialization. We provide the calculation of gradient norm (GN) and GradCosine (GC) for an initialized network in Algorithm \ref{alg1}.


\section{Neural Initialization Optimization}
\label{sec_NIO}

Based on the GradCosine, we propose our Neural Initialization Optimization (NIO) algorithm. Our algorithm follows \cite{zhu2021gradinit, dauphin2019metainit} to rectify the variance of the initialized parameters in a network via gradient descent. Concretely, we introduce a set of learnable scalar coefficients denoted as $M = \{\omega_1,\cdots,\omega_m\}$. The initial parameters $\theta_0=\{W_1,\cdots,W_m\}$ rectified by these coefficients can be formulated as $\theta_M=\{\omega_1W_1,\cdots,\omega_mW_m\}$. The NIO algorithm mainly solves the following constrained optimization problem:
\begin{equation}
	\begin{aligned}
		&\mathop{\text{maximize}}\limits_M \quad \text{GC}(S, \theta_M) + \text{GN}(S, \theta_M),\\
		&\text{subject to}  \quad \max_i(||g_{i}||_2) \leq \gamma,\ \ i = 1,...,B,
		\label{eq7}
	\end{aligned}
\end{equation}
where GC and GN refer to GradCosine and gradient norm, respectively, as calculated in Algorithm \ref{alg1}, and $B$ is the batchsize of the batch data $S$. With the maximal sample-wise gradient norm bounded by $\gamma$, we maximize both GradCosine and gradient norm. 

Although the problem in Eq. (\ref{eq7}) is differentiable and tractable, inferring GradCosine and gradient norm in the sample-wise manner can be very time-consuming especially for networks trained on large datasets such as ImageNet. 
Moreover, using gradient-based methods to solve Eq. (\ref{eq7}) requires deriving the second-order gradient with respect to the network parameters, which leads to unbearable memory consumption in the sample-wise manner. 
To this end, we introduce the batch-wise relaxation. 

\begin{algorithm}[t]
	\small
\caption{
Batch GradCosine (B-GC) and batch gradient norm (B-GN)
}
\label{alg2}
\begin{algorithmic}[1]
	\small
\REQUIRE Initialized network parameters $\theta_0$, a batch of samples $S = \{(x_i,y_i)\}_{i\in[B]}$, the number of sub-batches $D$, and overlap ratio $r$.
\FOR{$d=1$ to $D$}
\STATE Compute the batch-wise gradient: \\ $g_d \leftarrow \frac{1}{N}\sum_{j\in S_d} \nabla_\theta l(f_\theta(x_j), y_j)|_{\theta_0}$
\ENDFOR
\STATE Compute $\textbf{B-GN}(S,\theta_0)$ and $\textbf{B-GC}(S,\theta_0)$ with batch-wise gradients following Lines (4-6) in Algorithm \ref{alg1};
\STATE Output \textbf{B-GN} and \textbf{B-GC}.

\end{algorithmic}
\end{algorithm}

\subsection{Generalizing to Batch Settings}
\begin{figure}[t]
\vspace{2mm}
\begin{center}
\centerline{\includegraphics[width=\columnwidth]{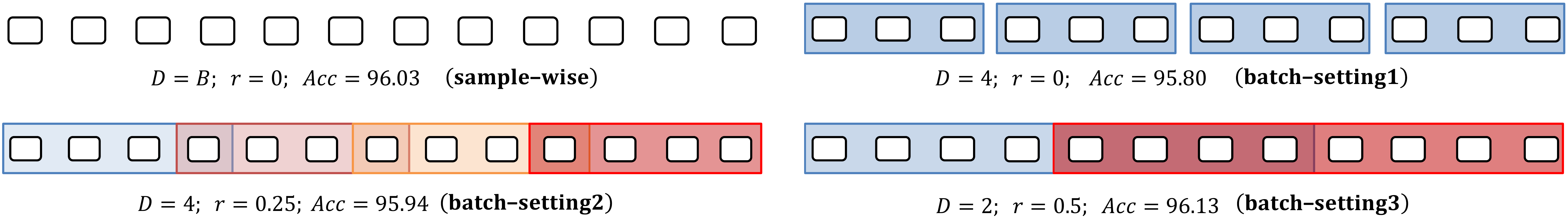}}
\vspace{-1mm}
\caption{An illustration of sample-wise inputs and batch-wise inputs with different batch settings. The example experiment is conducted on CIFAR-10 with ResNet-110.}
\label{fig2}
\end{center}
\vskip -0.1in
\end{figure}

To reduce the time and space complexity of gradient calculation, we further generalize GC and GN from the sample-wise manner into a more efficient batch setting. Specifically, for a mini-batch of training samples $S = \{(x_i,y_i)\}_{i\in[B]}$, we split it into $D$ sub-batches $S_1,...,S_d,...,S_D$ with an overlap ratio of $r$ that is used to relieve the violation of the sample-wise setting. The number of samples in each sub-batch is $N=\lceil \frac{B}{D-r} \rceil$, and each sub-batch can be denoted as:
\begin{equation}
    S_d = \{(x_i,y_i)\}_{i=N(d-1)(1-r)+1,...,N(d-1)(1-r)+N},
    \label{eq8}
\end{equation}
 where $d=1,...,D$. The sample-wise input can be regarded as the special case of the batch split when setting $D=B, r=0$. We further illustrate some split examples in Figure \ref{fig2}. We empirically find that making the sub-batches overlapped with each other stabilize the optimization and results in better initialization. As shown in Figure \ref{fig2}, the test accuracies of the models trained using our initialization with different batch settings further confirm this practice.

The GradCosine and gradient norm under the batch setting are illustrated in Algorithm \ref{alg2}. The main difference is that, in the batch setting, the gradient is computed as the average of gradients in a sub-batch. Our optimization problem in the batch setting can be formulated as:
\begin{equation}
\begin{aligned}
    &\mathop{\text{maximize}}\limits_M \quad \text{B-GC}(S, \theta_M) + \text{B-GN}(S, \theta_M),\\
    &\text{subject to}  \quad \max_i(||g_{i}||_2) \leq \gamma,\ \ i = 1,...,D,
    \label{eq9}
\end{aligned}
\end{equation}
where B-GC and B-GN refer to the batch-wise GradCosine and gradient norm as calculated in Algorithm \ref{alg2}.

\subsection{Main Algorithm}

\begin{algorithm}[t]
\caption{
Neural Initialization Optimization
}
\label{alg3}
\begin{algorithmic}[1]
	\small
\REQUIRE Initialized parameters $\theta_0\in\mathbb{R}^m$, learning rate $\tau$ for scale coefficients $M$, upper bound of gradient norm $\gamma$, lower bound of the scale coefficients $\underline\alpha=0.01$, total iterations $T$, batch size $B$, the number of sub-batches $D$, and overlap ratio $r$.
\STATE $M^{(1)}\leftarrow \textbf{1} $, where $\textbf{1}$ denotes the all-ones vector in $\mathbb{R}^m$;
\FOR{$t=1$ to $T$}
\STATE Sample $S_t$ from the training set;
\STATE Compute $g_i$, $\text{B-GC}(S_t, \theta_{M^{(t)}})$, and $\text{B-GN}(S_t, \theta_{M^{(t)}})$ by Algorithm \ref{alg2};
\STATE Compute the $g_{max}=\max_i(||g_{i}||_2)$, $i = 1,...,D$;
\IF{$g_{max}>\gamma$}
\STATE $M^{(t+1)} \leftarrow M^{(t)}-\tau \nabla_{M^{(t)}}\text{B-GN}$
\ELSE{}
\STATE $M^{(t+1)} \leftarrow M^{(t)}+\tau \nabla_{M^{(t)}}(\text{B-GC}+\text{B-GN})$
\ENDIF
\STATE Clamp $M^{(t+1)}$ using $\underline\alpha$;
\ENDFOR
\STATE  Output the rectified initialization parameters $\theta_M{^{(T)}}$.
\end{algorithmic}
\end{algorithm}

The final neural initialization optimization is illustrated in Algorithm \ref{alg3}. It rectifies the initialized parameters by gradient descent to solve the constrained optimization problem in Eq. (\ref{eq9}). 





Concretely, we iterate for $T$ iterations. At each iteration, a random batch data $S_t$ is sampled from the training set, and divided into sub-batches according to $D$ and $r$. We calculate B-GN and B-GC by Algorithm \ref{alg2}. If the maximal gradient norm is larger than the predefined upper bound $\gamma$, we minimize the averaged batch gradient norm to avoid gradient explosion at initialization. When the constraint is satisfied, we maximize the GradCosine (B-GC) and gradient norm (B-GN) objectives simultaneously, in order to minimize the upper bound of the quantity $\Theta$ in Eq. (\ref{relax_theta}), which intimately corresponds to not only a lower training and generalization error, but also a better training dynamic. 



\section{Experiments}

\subsection{Initialization for CNN}
\label{s61}
\textbf{Dataset and Architectures.}
We validate our method on three widely used datasets including CIFAR10/100 \cite{krizhevsky2009learning} and ImageNet \cite{deng2009imagenet}. We select three kinds of convolution neural networks including ResNet \cite{he2016deep}, DenseNet \cite{huang2017densely}, and WideResNet \cite{zagoruyko2016wide}.
On CIFAR10/100, we adopt ResNet110, DenseNet100, and the 28-layer Wide ResNet with Widen Factor 10 (WRN-28-10) as three main architectures for evaluation. To further show that our initialization method helps training dynamic for better training stability, we also conduct experiments on the same architectures but remove the batch normalization (BN) \cite{ioffe2015batch} layers. Moreover, to illustrate that our method can be extensible to large-scale benchmarks, we test our proposed NIO on ImageNet using ResNet-50. 
The detailed settings for different architectures and datasets are described in Appendix \ref{ap2}.


\begin{table}
	\small
	\vspace{-1mm}
	\begin{minipage}[th!]{\textwidth}
		\begin{minipage}[t]{0.48\textwidth}
			\renewcommand\arraystretch{0.9}
			\setlength{\tabcolsep}{2.8pt}
			\centering
			\caption{Test accuracies of three architectures on CIFAR-10. Best
				results are marked in bold.}
\begin{tabular}{l|ccc}
	\toprule
	Model & ResNet-110 & DenseNet-100 & WRN-28-10 \\
	\midrule
	& \multicolumn{3}{c}{w/ BN} \\
	\midrule
	Kaiming    & 95.53$\pm$ 0.19& 95.75$\pm$ 0.13& 97.27$\pm$ 0.27 \\
	Warmup     & 95.56$\pm$ 0.12& 95.73$\pm$ 0.27& 97.30$\pm$ 0.18\\
	Metainit   & 95.45$\pm$ 0.33& 95.75$\pm$ 0.11& 97.26$\pm$ 0.17 \\
	Gradinit   & 95.81$\pm$ 0.29& 95.77$\pm$ 0.22& 97.34$\pm$ 0.15 \\
	NIO  & \textbf{96.13}$\pm$ 0.16& \textbf{96.07}$\pm$ 0.18& \textbf{97.50}$\pm$ 0.21\\
	\midrule
	& \multicolumn{3}{c}{w/o BN} \\
	\midrule
	Kaiming    & 94.83$\pm$ 0.24& 94.78$\pm$ 0.16&  97.15$\pm$ 0.33 \\
	Warmup     & 94.75$\pm$ 0.21& 94.86$\pm$ 0.15&  97.20$\pm$ 0.18\\
	Metainit   & 94.79$\pm$ 0.15& 95.15$\pm$ 0.19&  97.27$\pm$ 0.25 \\
	Gradinit   & 95.12$\pm$ 0.16& 95.31$\pm$ 0.37& 97.12$\pm$ 0.13        \\
	NIO       & \textbf{95.27}$\pm$ 0.19& \textbf{95.62}$\pm$ 0.18& \textbf{97.35}$\pm$ 0.19\\
	\bottomrule
\end{tabular}
			\vspace{2mm}
			\label{tab1}
		\end{minipage}
		\quad \
		\begin{minipage}[t]{0.48\textwidth}
			\renewcommand\arraystretch{0.9}
			\setlength{\tabcolsep}{2.8pt}
			\centering
			\caption{Test accuracies of three architectures on CIFAR-100. Best
				results are marked in bold.}
\begin{tabular}{l|ccc}
	\toprule
	Model & ResNet-110 & DenseNet-100 & WRN-28-10 \\
	\midrule
	& \multicolumn{3}{c}{w/ BN} \\
	\midrule
	Kaiming    & 74.54$\pm$ 0.21& 76.58$\pm$ 0.23& 81.40$\pm$ 0.28 \\
	Warmup     & 74.63$\pm$ 0.33& 76.60$\pm$ 0.32& 81.45$\pm$ 0.25\\
	Metainit   & 74.32$\pm$ 0.26& 76.23$\pm$ 0.28& 81.46$\pm$ 0.20 \\
	Gradinit   & 75.40$\pm$ 0.17& 76.14$\pm$ 0.21& 81.35$\pm$ 0.31         \\
	NIO       & \textbf{75.72}$\pm$ 0.15& \textbf{76.86}$\pm$ 0.26& \textbf{81.83}$\pm$ 0.20\\
	\midrule
	& \multicolumn{3}{c}{w/o BN} \\
	\midrule
	Kaiming    & 73.03$\pm$ 0.23& 71.27$\pm$ 0.25& 79.28$\pm$ 0.26 \\
	Warmup     & 73.10$\pm$ 0.14&  71.50$\pm$ 0.24& 79.58$\pm$ 0.13\\
	Metainit   & 72.60$\pm$ 0.17&  71.68$\pm$ 0.37& 79.15$\pm$ 0.24 \\
	Gradinit   & 72.06$\pm$ 0.31& 71.33$\pm$ 0.21& 79.64$\pm$ 0.23         \\
	NIO       & \textbf{73.17}$\pm$ 0.15& \textbf{72.79}$\pm$ 0.22& \textbf{79.76}$\pm$ 0.26\\
	\bottomrule
\end{tabular}
			\vspace{2.5mm}
			\label{tab2}
		\end{minipage}
	\end{minipage}
	\vspace{-3mm}
\end{table}

\textbf{Experiment Results on CIFAR-10/100.}
We select four different initialization methods for comparison: (1) Kaiming Initialization \cite{he2015delving}; (2) First train the network for one epoch with a linear warmup learning rate, denoted as Warmup; (3) MetaInit \cite{dauphin2019metainit}; and (4) GradInit \cite{zhu2021gradinit}. (3) and (4) are learning-based initialization as ours. We re-implement their methods using the code provided in \cite{zhu2021gradinit}. For MetaInit, GradInit, and our proposed NIO, the initialization is rectified based on the Kaiming initialized parameters. After initialization, we train these models for 500 epochs with the same training setting. Each model is trained four times with different seeds. We report the average and standard deviation numbers of the accuracies on the test set.


As shown in Table \ref{tab1} and \ref{tab2}, compared with the Kaiming initialization, we observe that MetaInit and GradInit do not achieve improvement for some cases. As a comparison, our proposed NIO consistently improves upon the Kaiming initialization on both CIFAR-10 and CIFAR-100 with and without BN. Especially for ResNet-110-BN and DenseNet-100-noBN, the performance improvements upon the Kaiming initialization are more than 0.5\% on CIFAR-10 and 1.0\% on CIFAR-100. Compared with GradInit, we also achieve better results by more than 1.0\% for ResNet-110-noBN and DenseNet-110-noBN on CIFAR-100. 

\begin{figure}[t]
\begin{center}
\centerline{\includegraphics[width=\columnwidth]{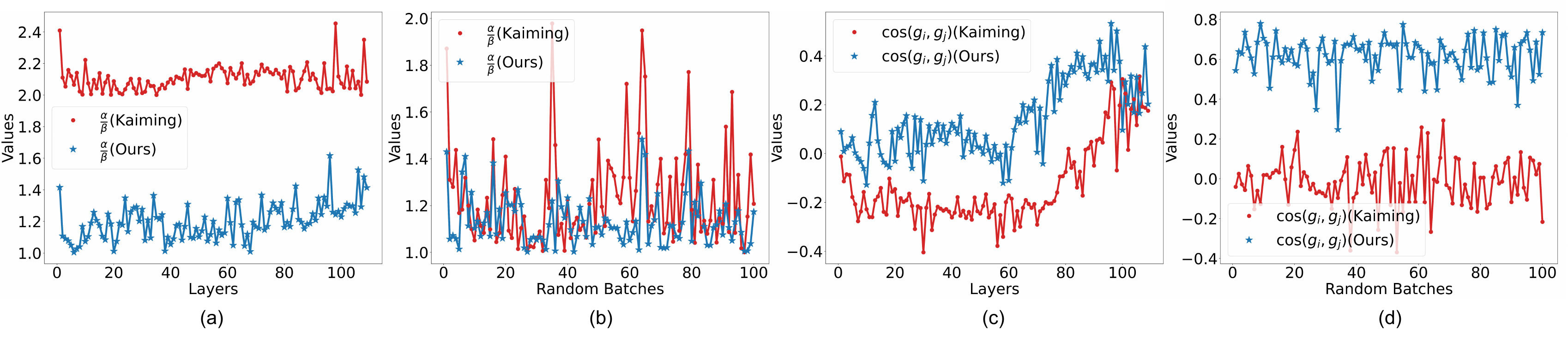}}
\vspace{-1mm}
\caption{Empirical comparisons between the Kaiming initialization (red) and our NIO (blue) using ResNet-110 on CIFAR-10. (a) and (b) show the ratio of the maximal sample-wise gradient norm to the minimal one. (c) and (d) depict the averaged pair-wise cosine similarities between sample-wise gradients. (a) and (c) are calculated for each layer with a random batch. (b) and (d) are calculated for all parameters using 100 randomly sampled batches. Zoom in for a better view.}
	
\label{fig3}
\end{center}
\vskip -0.2in

\end{figure}

We also make some empirical comparisons between the Kaiming initialization and our NIO with ResNet-110 on CIFAR-10. As shown in Figure \ref{fig3}, compared with the Kaiming initialization, NIO enjoys significantly smaller ratios of the maximal gradient norm to the minimal one, and larger cosine similarities of sample-wise gradients. Therefore, after NIO, the gradient norm ratio is reduced towards 1.0, while the cosine similarity is improved towards 1.0, which indicates that the quantity in Eq. (\ref{eq2}) is minimized, and the sample-wise gradients are close with respect to length and direction.




\textbf{Ablation Studies on Batch Setting.} We further test the sensitivity of our method to the batch setting hyper-parameters in Eq. (\ref{eq8}), the number of sub-batches $D$ and overlap ratio $r$. We show the model performances and the time and space complexities of NIO with ResNet-110 on CIFAR-100 using different $D$ and $r$ in Tabel \ref{ablation}. The performances do not vary significantly with the change of $D$ and $r$. The worst accuracy (when $D=2$ and $r=0$ ) 75.56\% in Table \ref{ablation} is still higher than the Kaiming initialization result 74.54\% by 1\%. As $D$ increases, the time and memory consumption of NIO also grows, because more batch gradients need to be  calculated by back propagation. When $D$ is small, a larger $r$ should be adopted to relieve the violation of the sample-wise hypothesis.



\textbf{Experiment Results on ImageNet.} To verify that our method can be extensible to large-scale task, we further evaluate our approach on the ImageNet dataset. Using different initialization methods, we train ResNet-50 for 100 epochs with a batchsize of 256. Alll models are trained under the same training setting. As shown in Table \ref{tab3}, we improve upon the Kaiming initialization by 0.3\%. We also re-implement GradInit using the default configurations in their paper \cite{zhu2021gradinit}, and have a performance close to the Kaiming initialization baseline.  Moreover, NIO takes the shortest initialization time compared with the other two learning-based initialization methods, MetaInit and GradInit. It is because only 100 iterations are needed for NIO in our implementation, while MetaInit and GradInit require 1000 and 2000 iterations, respectively, as their default settings \cite{dauphin2019metainit,zhu2021gradinit}. Compared with the training time, the initialization time of NIO is negligible.



\subsection{Initialization for Transformer}
\label{s62}
Vision Transformer architectures have been popular and proven to be effective for many vision tasks \cite{dosovitskiy2020image,touvron2021training,liu2021swin,xu2021vitae}. But Transformer architectures suffer from unstable training due to the large gradient variance \cite{liu2019variance,zhuang2020adabelief} in the Adam optimizer \cite{kingma2015adam}, so have to resort to the warmup trick, which adopts a small learning rate at initialization to stabilize the training process. In order to test the effectiveness of NIO to improve the training stability, we perform NIO for Swin-Transformer \cite{liu2021swin} with and without warmup on ImageNet. Detailed training and initialization settings are described in Appendix \ref{ap2}. As shown in Table \ref{tab4}, when warmup is not enabled for training, both Kaiming and Truncated normal initialized models cannot converge. As a comparison, the model with NIO achieves a top-1 accuracy of 80.9\%, which is very close to the standard training result with warmup. These results indicate that NIO is able to produce a better initialization agnostic of architecture and dataset. 

\begin{table}[t]
	\small
	\begin{minipage}[th!]{\textwidth}
		\begin{minipage}[t]{0.51\textwidth}
			\renewcommand\arraystretch{1.1}
			\setlength{\tabcolsep}{3pt}
			\centering
			\caption{Model performance and complexity of ResNet-110 on CIFAR-100 with different batch settings. NIO is performed with a batchsize of 128. A smaller number of sub-batches $D$ leads to faster speed and less memory consumption, and almost does not harm the final performance. ``Time'' refers to the time consumed for each iteration in NIO.}
			\begin{tabular}{c|cccc|c|c}
				\hline
				\multirow{2}{*}{$D$}& \multicolumn{4}{c|}{Accuracy (\%)}& \multirow{2}{*}{Time} & \multirow{2}{*}{GPU Mem.} \\\cline{2-5}
				&$r$=0 & $r$=0.2 & $r$=0.4 & $r$=0.6  &  &  \\\hline
				2&  75.56 & 75.71 & \textbf{75.96} &75.72&\textbf{3.75}s &\textbf{5007}M  \\
				3& 75.60 & \textbf{76.00}  & 75.97 & 75.76 &4.07s&6001M      \\
				4& 75.58 & \textbf{76.03} & 75.71 & 75.88&4.68s&7067M \\\hline
			\end{tabular}
			
			\label{ablation}
		\end{minipage}
		\quad \
		\begin{minipage}[t]{0.51\textwidth}
			\renewcommand\arraystretch{0.87}
			\setlength{\tabcolsep}{5.0pt}
			\centering
			\caption{Top-1 accuracy of ResNet-50 on ImageNet. The results of FixUp and MetaInit are taken from their papers \cite{zhang2019fixup,dauphin2019metainit}, and the other methods are re-implemented by us under the same training setting. FixUp removes the BN layers as their default and the other models are trained with BN layers. }
		\label{tab3}
		\begin{tabular}{l|c|c}
			\toprule
			Method & Accuracy (\%)  & Init. / Train time \tablefootnote{``Train time'' is tested on an NVIDIA A100 server with a batchsize of 256 among 8 GPUs. ``Init. time'' of MetaInit and GradInit is tested for 1000 and 2000 iterations, respectively, according to the default configurations in their papers \cite{dauphin2019metainit,zhu2021gradinit}. NIO is performed for 100 iterations so enjoys a fast initialization process.}\\
			\midrule
			Kaiming \cite{he2015delving}  & 76.43$\pm$0.05 &  - / 8.5h \\
			FixUp \cite{zhang2019fixup} & 76.0 &  - / 8.5h \\
			MetaInit \cite{dauphin2019metainit} & 76.0  & 0.81h  / 8.5h \\
			GradInit  \cite{zhu2021gradinit} & 76.50$\pm$0.05 & 0.21h / 8.5h\\
			NIO (ours) & \textbf{76.71}$\pm$0.07 & 0.03h / 8.5h \\
			\bottomrule
		\end{tabular}
	\end{minipage}
\end{minipage}
\end{table}

\begin{table}[t]
\vspace{1mm}
	\caption{Top-1 accuracy of Swin-Transformer (tiny) on ImageNet with and without warmup. ``fail'' represents that the model cannot converge. Warmup takes up the first 20 epochs when enabled.}
\vspace{-2mm}	
	\label{tab4}
	\begin{center}
		\begin{small}
			\begin{tabular}{lcccc}
				\toprule
				& Kaiming & TruncNormal & GradInit\tablefootnote{GradInit did not perform on SwinTransformer in their paper. We adopt their initialization settings for ResNet-50 to perform GradInit on SwinTransformer and then train the model.} & NIO (ours) \\
				\midrule
				w/ warmup  &79.4 & 81.3 & 80.4 & \textbf{81.3} \\
				w/o warmup & fail & fail & 79.9 & \textbf{80.9} \\
				\bottomrule
			\end{tabular}
		\end{small}
	\end{center}
	\vspace{-3mm}
\end{table}

\subsection{Discussions}
\label{dicuss}

\textbf{Why not Gradient Variance?} As indicated by Eq. (\ref{eq2}) and Figure \ref{fig3}, the aim of NIO is to reduce the variability of sample-wise gradients.
But why do we not directly minimize the gradient variance or just the pair-wise Euclidean distances? We point out that gradient variance is highly sensitive to gradient norm. Directly minimizing gradient variance would lead to a trivial gradient norm near zero, which disables the training. So the ability of reducing gradient variance while keeping gradient norm at a proper range is one crucial property of our proposed NIO.

\textbf{Limitations and Societal Impacts.} Similar to \cite{zhu2021gradinit}, the choice of $\gamma$ has not been determined from a theoretical point yet. Calculating GradCosine accurately requires much memory consumption for large models, so it needs the batch setting relaxation. Besides, performing NIO without a dataset will be also a promising breakthrough that deserves future exploration. Our study is general and not concerned with malicious practices with respect to privacy, public health, fairness, \emph{etc}. The proposed algorithm is lightweight and does not bring much environment burden.

\section{Conclusion}

In this paper, we propose a differentiable quantity with theoretical insight to evaluate the initial state of a neural network. Based on our sample-wise landscape analysis, we show that maximizing the quantity with the maximal gradient norm upper bounded is able to improve both training dynamic and model performance. Accordingly, we develop an initialization optimization algorithm to rectify the initial parameters. Experimental results show that our method is able to automatically produce a better initialization for variant architectures to improve the performance on multiple datasets with negligible cost. It also helps a Transformer architecture train stably even without warmup.



\begin{ack}
	Z. Lin was supported by the NSF China (No.s 62276004 and 61731018), NSFC Tianyuan Fund for Mathematics (No. 12026606), and Project 2020BD006 supported by PKU-Baidu Fund.
\end{ack}


\small
\bibliography{example_paper}

\begin{thebibliography}{10}\itemsep=-1pt

\bibitem{abdelfattah2020zero}
M.~S. Abdelfattah, A.~Mehrotra, {\L}.~Dudziak, and N.~D. Lane.
\newblock Zero-cost proxies for lightweight nas.
\newblock In {\em ICLR}, 2021.

\bibitem{allen2019convergence}
Z.~Allen-Zhu, Y.~Li, and Z.~Song.
\newblock A convergence theory for deep learning via over-parameterization.
\newblock In {\em ICML}, pages 242--252. PMLR, 2019.

\bibitem{bergstra2011algorithms}
J.~Bergstra, R.~Bardenet, Y.~Bengio, and B.~K{\'e}gl.
\newblock Algorithms for hyper-parameter optimization.
\newblock In {\em NeurIPS}, volume~24, 2011.

\bibitem{brutzkus2017globally}
A.~Brutzkus and A.~Globerson.
\newblock Globally optimal gradient descent for a convnet with gaussian inputs.
\newblock In {\em ICML}, pages 605--614. PMLR, 2017.

\bibitem{chen2018searching}
L.-C. Chen, M.~Collins, Y.~Zhu, G.~Papandreou, B.~Zoph, F.~Schroff, H.~Adam,
  and J.~Shlens.
\newblock Searching for efficient multi-scale architectures for dense image
  prediction.
\newblock In {\em NeurIPS}, volume~31, 2018.

\bibitem{chen2021neural}
W.~Chen, X.~Gong, and Z.~Wang.
\newblock Neural architecture search on imagenet in four gpu hours: A
  theoretically inspired perspective.
\newblock In {\em ICLR}, 2020.

\bibitem{das2021data}
D.~Das, Y.~Bhalgat, and F.~Porikli.
\newblock Data-driven weight initialization with sylvester solvers.
\newblock {\em arXiv preprint arXiv:2105.10335}, 2021.

\bibitem{dauphin2019metainit}
Y.~N. Dauphin and S.~Schoenholz.
\newblock Metainit: Initializing learning by learning to initialize.
\newblock In {\em NeurIPS}, volume~32, 2019.

\bibitem{deng2009imagenet}
J.~Deng, W.~Dong, R.~Socher, L.-J. Li, K.~Li, and L.~Fei-Fei.
\newblock Imagenet: A large-scale hierarchical image database.
\newblock In {\em CVPR}, pages 248--255, 2009.

\bibitem{devries2017improved}
T.~DeVries and G.~W. Taylor.
\newblock Improved regularization of convolutional neural networks with cutout.
\newblock {\em arXiv preprint arXiv:1708.04552}, 2017.

\bibitem{dosovitskiy2020image}
A.~Dosovitskiy, L.~Beyer, A.~Kolesnikov, D.~Weissenborn, X.~Zhai,
  T.~Unterthiner, M.~Dehghani, M.~Minderer, G.~Heigold, S.~Gelly, et~al.
\newblock An image is worth 16x16 words: Transformers for image recognition at
  scale.
\newblock In {\em ICLR}, 2020.

\bibitem{du2018gradient}
S.~Du, J.~Lee, Y.~Tian, A.~Singh, and B.~Poczos.
\newblock Gradient descent learns one-hidden-layer cnn: Don’t be afraid of
  spurious local minima.
\newblock In {\em ICML}, pages 1339--1348. PMLR, 2018.

\bibitem{feurer2019hyperparameter}
M.~Feurer and F.~Hutter.
\newblock Hyperparameter optimization.
\newblock In {\em Automated machine learning}, pages 3--33. Springer, Cham,
  2019.

\bibitem{ge2017learning}
R.~Ge, J.~D. Lee, and T.~Ma.
\newblock Learning one-hidden-layer neural networks with landscape design.
\newblock In {\em ICLR}, 2018.

\bibitem{gilmer2021loss}
J.~Gilmer, B.~Ghorbani, A.~Garg, S.~Kudugunta, B.~Neyshabur, D.~Cardoze,
  G.~Dahl, Z.~Nado, and O.~Firat.
\newblock A loss curvature perspective on training instability in deep
  learning.
\newblock {\em arXiv preprint arXiv:2110.04369}, 2021.

\bibitem{glorot2010understanding}
X.~Glorot and Y.~Bengio.
\newblock Understanding the difficulty of training deep feedforward neural
  networks.
\newblock In {\em AISTATS}, pages 249--256, 2010.

\bibitem{goyal2017accurate}
P.~Goyal, P.~Doll{\'a}r, R.~Girshick, P.~Noordhuis, L.~Wesolowski, A.~Kyrola,
  A.~Tulloch, Y.~Jia, and K.~He.
\newblock Accurate, large minibatch sgd: Training imagenet in 1 hour.
\newblock {\em arXiv preprint arXiv:1706.02677}, 2017.

\bibitem{he2015delving}
K.~He, X.~Zhang, S.~Ren, and J.~Sun.
\newblock Delving deep into rectifiers: Surpassing human-level performance on
  imagenet classification.
\newblock In {\em ICCV}, pages 1026--1034, 2015.

\bibitem{he2016deep}
K.~He, X.~Zhang, S.~Ren, and J.~Sun.
\newblock Deep residual learning for image recognition.
\newblock In {\em CVPR}, pages 770--778, 2016.

\bibitem{howard2017mobilenets}
A.~G. Howard, M.~Zhu, B.~Chen, D.~Kalenichenko, W.~Wang, T.~Weyand,
  M.~Andreetto, and H.~Adam.
\newblock Mobilenets: Efficient convolutional neural networks for mobile vision
  applications.
\newblock {\em arXiv preprint arXiv:1704.04861}, 2017.

\bibitem{huang2017densely}
G.~Huang, Z.~Liu, L.~Van Der~Maaten, and K.~Q. Weinberger.
\newblock Densely connected convolutional networks.
\newblock In {\em CVPR}, pages 4700--4708, 2017.

\bibitem{huang2020explicitly}
T.~Huang, S.~You, Y.~Yang, Z.~Tu, F.~Wang, C.~Qian, and C.~Zhang.
\newblock Explicitly learning topology for differentiable neural architecture
  search.
\newblock {\em arXiv preprint arXiv:2011.09300}, 2020.

\bibitem{huang2020improving}
X.~S. Huang, F.~Perez, J.~Ba, and M.~Volkovs.
\newblock Improving transformer optimization through better initialization.
\newblock In {\em ICML}, pages 4475--4483. PMLR, 2020.

\bibitem{ioffe2015batch}
S.~Ioffe and C.~Szegedy.
\newblock Batch normalization: Accelerating deep network training by reducing
  internal covariate shift.
\newblock In {\em ICML}, pages 448--456. PMLR, 2015.

\bibitem{jacot2018neural}
A.~Jacot, F.~Gabriel, and C.~Hongler.
\newblock Neural tangent kernel: Convergence and generalization in neural
  networks.
\newblock In {\em NeurIPS}, volume~31, 2018.

\bibitem{kingma2015adam}
D.~P. Kingma and J.~Ba.
\newblock Adam: A method for stochastic optimization.
\newblock In {\em ICLR}, 2015.

\bibitem{krizhevsky2009learning}
A.~Krizhevsky, G.~Hinton, et~al.
\newblock Learning multiple layers of features from tiny images.
\newblock 2009.

\bibitem{lee2018snip}
N.~Lee, T.~Ajanthan, and P.~H. Torr.
\newblock Snip: Single-shot network pruning based on connection sensitivity.
\newblock In {\em ICLR}, 2019.

\bibitem{li2018optimization}
H.~Li, Y.~Yang, D.~Chen, and Z.~Lin.
\newblock Optimization algorithm inspired deep neural network structure design.
\newblock In {\em ACML}, 2018.

\bibitem{li2017convergence}
Y.~Li and Y.~Yuan.
\newblock Convergence analysis of two-layer neural networks with relu
  activation.
\newblock In {\em NeurIPS}, volume~30, 2017.

\bibitem{liu2019variance}
L.~Liu, H.~Jiang, P.~He, W.~Chen, X.~Liu, J.~Gao, and J.~Han.
\newblock On the variance of the adaptive learning rate and beyond.
\newblock In {\em ICLR}, 2020.

\bibitem{liu2021swin}
Z.~Liu, Y.~Lin, Y.~Cao, H.~Hu, Y.~Wei, Z.~Zhang, S.~Lin, and B.~Guo.
\newblock Swin transformer: Hierarchical vision transformer using shifted
  windows.
\newblock In {\em ICCV}, pages 10012--10022, 2021.

\bibitem{loshchilov2016sgdr}
I.~Loshchilov and F.~Hutter.
\newblock Sgdr: Stochastic gradient descent with warm restarts.
\newblock In {\em ICLR}, 2017.

\bibitem{mellor2021neural}
J.~Mellor, J.~Turner, A.~Storkey, and E.~J. Crowley.
\newblock Neural architecture search without training.
\newblock In {\em ICML}, pages 7588--7598. PMLR, 2021.

\bibitem{meuleau2002ant}
N.~Meuleau and M.~Dorigo.
\newblock Ant colony optimization and stochastic gradient descent.
\newblock {\em Artificial life}, 8(2):103--121, 2002.

\bibitem{mishkin2015all}
D.~Mishkin and J.~Matas.
\newblock All you need is a good init.
\newblock {\em arXiv preprint arXiv:1511.06422}, 2015.

\bibitem{oymak2019overparameterized}
S.~Oymak and M.~Soltanolkotabi.
\newblock Overparameterized nonlinear learning: Gradient descent takes the
  shortest path?
\newblock In {\em ICML}, pages 4951--4960. PMLR, 2019.

\bibitem{real2019regularized}
E.~Real, A.~Aggarwal, Y.~Huang, and Q.~V. Le.
\newblock Regularized evolution for image classifier architecture search.
\newblock In {\em AAAI}, volume~33, pages 4780--4789, 2019.

\bibitem{ruder2016overview}
S.~Ruder.
\newblock An overview of gradient descent optimization algorithms.
\newblock {\em arXiv preprint arXiv:1609.04747}, 2016.

\bibitem{saxe2013exact}
A.~M. Saxe, J.~L. McClelland, and S.~Ganguli.
\newblock Exact solutions to the nonlinear dynamics of learning in deep linear
  neural networks.
\newblock {\em arXiv preprint arXiv:1312.6120}, 2013.

\bibitem{shu2021nasi}
Y.~Shu, S.~Cai, Z.~Dai, B.~C. Ooi, and B.~K.~H. Low.
\newblock Nasi: Label-and data-agnostic neural architecture search at
  initialization.
\newblock In {\em ICLR}, 2022.

\bibitem{simon2021neural}
J.~B. Simon, M.~Dickens, and M.~R. DeWeese.
\newblock Neural tangent kernel eigenvalues accurately predict generalization.
\newblock {\em arXiv preprint arXiv:2110.03922}, 2021.

\bibitem{soltanolkotabi2017learning}
M.~Soltanolkotabi.
\newblock Learning relus via gradient descent.
\newblock In {\em NeurIPS}, volume~30, 2017.

\bibitem{tanaka2020pruning}
H.~Tanaka, D.~Kunin, D.~L. Yamins, and S.~Ganguli.
\newblock Pruning neural networks without any data by iteratively conserving
  synaptic flow.
\newblock In {\em NeurIPS}, volume~33, pages 6377--6389, 2020.

\bibitem{theis2018faster}
L.~Theis, I.~Korshunova, A.~Tejani, and F.~Husz{\'a}r.
\newblock Faster gaze prediction with dense networks and fisher pruning.
\newblock {\em arXiv preprint arXiv:1801.05787}, 2018.

\bibitem{touvron2021training}
H.~Touvron, M.~Cord, M.~Douze, F.~Massa, A.~Sablayrolles, and H.~J{\'e}gou.
\newblock Training data-efficient image transformers \& distillation through
  attention.
\newblock In {\em ICML}, pages 10347--10357. PMLR, 2021.

\bibitem{turner2019blockswap}
J.~Turner, E.~J. Crowley, M.~O'Boyle, A.~Storkey, and G.~Gray.
\newblock Blockswap: Fisher-guided block substitution for network compression
  on a budget.
\newblock In {\em ICLR}, 2020.

\bibitem{turner2021neural}
J.~Turner, E.~J. Crowley, and M.~F. O'Boyle.
\newblock Neural architecture search as program transformation exploration.
\newblock In {\em Proceedings of the 26th ACM International Conference on
  Architectural Support for Programming Languages and Operating Systems}, pages
  915--927, 2021.

\bibitem{vaswani2017attention}
A.~Vaswani, N.~Shazeer, N.~Parmar, J.~Uszkoreit, L.~Jones, A.~N. Gomez,
  {\L}.~Kaiser, and I.~Polosukhin.
\newblock Attention is all you need.
\newblock In {\em NeurIPS}, pages 5998--6008, 2017.

\bibitem{wang2020picking}
C.~Wang, G.~Zhang, and R.~Grosse.
\newblock Picking winning tickets before training by preserving gradient flow.
\newblock In {\em ICLR}, 2020.

\bibitem{xiao2020disentangling}
L.~Xiao, J.~Pennington, and S.~Schoenholz.
\newblock Disentangling trainability and generalization in deep neural
  networks.
\newblock In {\em ICML}, pages 10462--10472. PMLR, 2020.

\bibitem{xiong2020layer}
R.~Xiong, Y.~Yang, D.~He, K.~Zheng, S.~Zheng, C.~Xing, H.~Zhang, Y.~Lan,
  L.~Wang, and T.~Liu.
\newblock On layer normalization in the transformer architecture.
\newblock In {\em ICML}, pages 10524--10533. PMLR, 2020.

\bibitem{xu2021vitae}
Y.~Xu, Q.~Zhang, J.~Zhang, and D.~Tao.
\newblock Vitae: Vision transformer advanced by exploring intrinsic inductive
  bias.
\newblock In {\em NeurIPS}, volume~34, 2021.

\bibitem{yang2020ista}
Y.~Yang, H.~Li, S.~You, F.~Wang, C.~Qian, and Z.~Lin.
\newblock Ista-nas: Efficient and consistent neural architecture search by
  sparse coding.
\newblock In {\em NeurIPS}, 2020.

\bibitem{yang2021towards}
Y.~Yang, S.~You, H.~Li, F.~Wang, C.~Qian, and Z.~Lin.
\newblock Towards improving the consistency, efficiency, and flexibility of
  differentiable neural architecture search.
\newblock In {\em CVPR}, 2021.

\bibitem{zagoruyko2016wide}
S.~Zagoruyko and N.~Komodakis.
\newblock Wide residual networks.
\newblock {\em arXiv preprint arXiv:1605.07146}, 2016.

\bibitem{zhang2017mixup}
H.~Zhang, M.~Cisse, Y.~N. Dauphin, and D.~Lopez-Paz.
\newblock mixup: Beyond empirical risk minimization.
\newblock In {\em ICLR}, 2018.

\bibitem{zhang2019fixup}
H.~Zhang, Y.~N. Dauphin, and T.~Ma.
\newblock Residual learning without normalization via better initialization.
\newblock In {\em ICLR}, 2019.

\bibitem{zhang2021gradsign}
Z.~Zhang and Z.~Jia.
\newblock Gradsign: Model performance inference with theoretical insights.
\newblock In {\em ICLR}, 2022.

\bibitem{zhu2021gradinit}
C.~Zhu, R.~Ni, Z.~Xu, K.~Kong, W.~R. Huang, and T.~Goldstein.
\newblock Gradinit: Learning to initialize neural networks for stable and
  efficient training.
\newblock In {\em NeurIPS}, volume~34, 2021.

\bibitem{zhuang2020adabelief}
J.~Zhuang, T.~Tang, Y.~Ding, S.~C. Tatikonda, N.~Dvornek, X.~Papademetris, and
  J.~Duncan.
\newblock Adabelief optimizer: Adapting stepsizes by the belief in observed
  gradients.
\newblock In {\em NeurIPS}, volume~33, pages 18795--18806, 2020.

\bibitem{zoph2016neural}
B.~Zoph and Q.~V. Le.
\newblock Neural architecture search with reinforcement learning.
\newblock In {\em ICLR}, 2017.

\end{thebibliography}
\bibliographystyle{ieee}

\normalsize
\section*{Checklist}

\begin{enumerate}

	\item For all authors...
	\begin{enumerate}
		\item Do the main claims made in the abstract and introduction accurately reflect the paper's contributions and scope?
		\answerYes{See the last three paragraphs in Section \ref{intro}.}
		\item Did you describe the limitations of your work?
		\answerYes{See Section \ref{dicuss}.}
		\item Did you discuss any potential negative societal impacts of your work?
		\answerYes{See Section \ref{dicuss}.}
		\item Have you read the ethics review guidelines and ensured that your paper conforms to them?
		\answerYes{We are ensured.}
	\end{enumerate}

	\item If you are including theoretical results...
	\begin{enumerate}
		\item Did you state the full set of assumptions of all theoretical results?
		\answerYes{See Section \ref{theory}.}
		\item Did you include complete proofs of all theoretical results?
		\answerYes{See Appendix A.}
	\end{enumerate}

	\item If you ran experiments...
	\begin{enumerate}
		\item Did you include the code, data, and instructions needed to reproduce the main experimental results (either in the supplemental material or as a URL)?
		\answerYes{See supplementary material.}
		\item Did you specify all the training details (e.g., data splits, hyperparameters, how they were chosen)?
		\answerYes{See Appendix B.}
		\item Did you report error bars (e.g., with respect to the random seed after running experiments multiple times)?
		\answerYes{See Table \ref{tab1} and \ref{tab2}.}
		\item Did you include the total amount of compute and the type of resources used (e.g., type of GPUs, internal cluster, or cloud provider)?
		\answerYes{See Table \ref{tab3}.}
	\end{enumerate}

	\item If you are using existing assets (e.g., code, data, models) or curating/releasing new assets...
	\begin{enumerate}
		\item If your work uses existing assets, did you cite the creators?
		\answerYes{We cite the corresponding papers.}
		\item Did you mention the license of the assets?
		\answerNA{}
		\item Did you include any new assets either in the supplemental material or as a URL?
		\answerNA{}
		\item Did you discuss whether and how consent was obtained from people whose data you're using/curating?
		\answerNA{}
		\item Did you discuss whether the data you are using/curating contains personally identifiable information or offensive content?
		\answerNA{}
	\end{enumerate}

	\item If you used crowdsourcing or conducted research with human subjects...
	\begin{enumerate}
		\item Did you include the full text of instructions given to participants and screenshots, if applicable?
		\answerNA{}
		\item Did you describe any potential participant risks, with links to Institutional Review Board (IRB) approvals, if applicable?
		\answerNA{}
		\item Did you include the estimated hourly wage paid to participants and the total amount spent on participant compensation?
		\answerNA{}
	\end{enumerate}

\end{enumerate}

\newpage
\appendix
\onecolumn

\begin{center}
	{\bf\large Towards Theoretically Inspired Neural Initialization Optimization\\(Appendix)}
\end{center}

\section{Proofs of Theoretical Results}
\label{ap1}


\subsection{Lemma 1 Proof}
\label{ap11}
Please refer to \cite{zhang2021gradsign} for the proof.

\subsection{Theorem 2 Proof}
\label{ap12}

Let $\theta^*$ denote an overall local optimum of the training loss $\mathcal{L}=\frac{1}{n}\sum^n_il(f_{\theta}(x_i),y_i)$ via gradient descent from the initialization $\theta_0$.
Since we have assumed that $\forall k\in [m], i\in [n], [\nabla^2l(f_{\theta}(x_i), y_i)]_{k,k}\leq\mathcal{H}$, 
through the expansion of $l(f_{\theta^*}(x_i),y_i)$ at the sample optimum $\theta^*_i$, we have:
\begin{equation}
    l(f_{\theta^*}(x_i),y_i) \leq l(f_{\theta^*_i}(x_i),y_i) + \nabla_\theta l(f_{\theta^*_i}(x_i),y_i)^T(\theta^*-\theta^*_i) + \frac{\mathcal{H}}{2}||\theta^*-\theta^*_i||^2_2.
    \label{eq14}
\end{equation}
At the sample optimum $\theta^*_i$, we have $l(f_{\theta^*_i}(x_i),y_i)=0$ and $\nabla_\theta l(f_{\theta^*_i}(x_i),y_i)=\mathbf{0}$. Then the above inequality can be rewrote as:
\begin{equation}
    l(f_{\theta^*}(x_i),y_i) \leq \frac{\mathcal{H}}{2}||\theta^*-\theta^*_i||^2_2,
    \label{eq15}
\end{equation}
and thus we have $\mathcal{L} \leq \frac{\mathcal{H}}{2n}\sum^n_i||\theta^*-\theta^*_i||^2_2$. Due to the assumption that the sample-wise optimization landscapes are nearly convex and semi-smooth at the neighborhood of the initialization $\theta_0$, the overall optimization landscape is also convex and semi-smooth, and the overall local optimum $\theta^*$ lies in the convex hull of the sample-wise optima $\theta^*_i, \forall i \in [1,n]$. Then we have:
\begin{equation}
    ||\theta^*-\theta^*_i||^2_2 \leq \sum_j ||\theta_j^*-\theta^*_i||^2_2,
    \label{eq16}
\end{equation}
and
\begin{align}
    \mathcal{L} \leq \frac{\mathcal{H}}{2n}\sum_{i,j}||\theta_i^*-\theta^*_j||^2_2.
\end{align}
We use $\alpha$ and $\beta$ to denote the maximal and minimal $\ell_2$-norms of the sample-wise optimization paths, respectively, \emph{i.e.,} $\alpha=\max(||\theta^*_i-\theta_0||_2), \beta=\min(||\theta^*_i-\theta_0||_2), \forall i \in [1,n]$.
We further denote the normalized sample-wise optimization paths as $\mathbf{p}_i = \frac{\theta^*_i-\theta_0}{\alpha}$, $0<||\mathbf{p}_i||_2\le1$. Then we have:
\allowdisplaybreaks[4]
\begin{align}
		\mathcal{L} & \leq \frac{\mathcal{H}}{2n}\sum_{i,j}||\theta_i^*-\theta^*_j||^2_2 \notag \\
		& = \frac{\mathcal{H}}{2n}\sum_{i,j}||\theta_i^*-\theta_0-(\theta^*_j-\theta_0)||^2_2  \notag \\
		& = \frac{\mathcal{H}\alpha^2}{2n}\sum_{i,j}\frac{||\theta_i^*-\theta_0-(\theta^*_j-\theta_0)||^2_2}{\alpha^2} \notag \\
		& = \frac{\mathcal{H}\alpha^2}{2n}\sum_{i,j}||\mathbf{p}_i-\mathbf{p}_j||_2^2 \label{theorem1}\\
		& \le \frac{\mathcal{H}\alpha^2}{2n}\sum_{i,j}\frac{||\mathbf{p}_i||_2^2+||\mathbf{p}_j||_2^2-2\mathbf{p}^T_i\mathbf{p}_j}{||\mathbf{p}_i||_2\cdot||\mathbf{p}_j||_2} \notag \\
		&= \frac{\mathcal{H}\alpha^2}{2n}\sum_{i,j}\left(\frac{||\mathbf{p}_i||_2}{||\mathbf{p}_j||_2} +\frac{||\mathbf{p}_j||_2}{||\mathbf{p}_i||_2} -2 \frac{\mathbf{p}^T_i\mathbf{p}_j}{||\mathbf{p}_i||_2\cdot||\mathbf{p}_j||_2}\right) \notag \\
		&\overset{a}{\le}  \frac{\mathcal{H}\alpha^2}{2n}\sum_{i,j}\left(2\frac{\alpha}{\beta} - 2 \cos\angle(\mathbf{p}_i,\mathbf{p}_j)\right) \notag \\
		&= \frac{\mathcal{H}\alpha^2}{n}\sum_{i,j}\left(\frac{\alpha}{\beta}-\cos\angle(\theta^*_i-\theta_0,\theta^*_j-\theta_0)\right)	\notag \\	&=\Theta_{S,l}(f_{\theta_0}(\cdot)),\notag 
\end{align}

where $\overset{a}{\le}$ holds because for any pair of $(i,j)$, $i,j \in [1,n]$, we have $\frac{||\mathbf{p}_i||_2}{||\mathbf{p}_j||_2}=\frac{||\theta^*_i-\theta_0||_2}{||\theta^*_j-\theta_0||_2}\le\frac{\alpha}{\beta}$.

\subsection{Theorem 3 Proof}
\label{ap13}
We use the population error $\mathbb{E}_{(x_u,y_u)\sim\mathcal{D}}[l(f_{\theta^*}(x_u), y_u)]$ to measure the generalization performance,
where $\mathcal{D}$ is the data distribution for both training and test data, and $(x_u,y_u)$ refers to the test sample.
With probability $1-\delta$, we have:
\begin{align}
    \mathbb{E}_{(x_u,y_u)\sim\mathcal{D}}[l(f_{\theta^*}(x_u), y_u)] 
    &\leq \frac{\mathcal{H}}{2}\mathbb{E}_{(x_u,y_u)\sim\mathcal{D}}[||\theta^*-\theta^*_u||^2_2] \label{eq21}  \\
    & \leq \frac{\mathcal{H}}{2n}\sum^n_i||\theta^*-\theta^*_i||^2_2 + \frac{\sigma}{\sqrt{n\delta}} \label{eq22} \\
    & \leq \Theta_{S,l}(f_{\theta_0}(\cdot))+\frac{\sigma}{\sqrt{n\delta}}, \label{eq23} 
\end{align}
where $n$ is the number of training samples and $\sigma$ is a positive constant such that $Var_{(x_u,y_u)\sim\mathcal D} [||\theta^*-\theta^*_u||^2_2]\le \sigma^2$. Eq. (\ref{eq21}) is based on Eq. (\ref{eq15}), Eq. (\ref{eq22}) holds following Chebyshev's inequality, and Eq. (\ref{eq23}) is derived from the result of Theorem 2.
With Theorem 2 and Theorem 3, we can conclude that $\Theta_{S,l}(f_{\theta_0}(\cdot))$ is the upper bound for both training and generalization error, and thus can be used as an indicator to evaluate the quality of the initialization $\theta_0$. 

\section{Experimental Details}
\label{ap2}
In this section, we provide more information about the implementation details, the architectures, the configurations of NIO, and the training details after NIO.
\subsection{Implementation Details}
\label{ap21}
\textbf{Selection of Hyper-parameters.}
The hypeparameters used in the NIO Algorithm \ref{alg3} are mainly selected according to the following principles: (1) For the number of sub-batches $D$ and overlap ratio $r$ in the batch-wise relaxation, while a smaller $D$ corresponds to a faster inference speed, it also means more violation of the sample-wise analysis. Therefore, the smaller $D$ is, the larger $r$ should be adopted to relieve the violation. Our ablation study also validates this practice. 
For all experiments, we set $D$ as 2 to obtain the fastest initialization speed and $r$ is correspondingly selected from $0.6\sim0.8$, proportional to the batch size $B$. (2) For the iteration steps $T$, we set it to a number that traverses all the training samples on CIFAR-10/100. For the large-scale dataset ImageNet, traversing all the training samples is time-consuming. So we set $T$ as 100 for experiments on ImageNet.
(3) For the upper bound of the gradient norm $\gamma$, we follow the choices suggested in \cite{zhu2021gradinit}. The learning rate $\tau$ of scale coefficients is set within the range from $10^{-3}$ to $1$. Networks with more parameters are initialized with a smaller learning rate $\tau$.


\subsection{On CIFAR-10/CIFAR-100}
\label{ap22}
\paragraph{Architecture Details.}
The architectures adopted include a ResNet-110 with an initial width of 16 and two convolution layers in each residual block; a 28-layer Wide ResNet with a widen factor of 10 (WRN-28-10); and a DenseNet-100. 
Following \cite{zhu2021gradinit}, for experiments without BN layers, we replace the BN layers with learnable bias parameters in the corresponding places to isolate the normalization effect from the affine function in BN.
When testing NIO on these architectures, we first adopt the non-zero Kaiming initialization to all convolution and FC layers, and then perform NIO upon the initialized parameters. 
These architecture settings remain unchanged for CIFAR-100 except that the channel size of the last linear layer is adjusted according to the number of categories.

\paragraph{NIO Configurations.}
For architectures with BN layers enabled, we select the learning rate $\tau$ from relatively larger values $\{3\times10^{-1}, 2.5\times10^{-1}, 2\times10^{-1}, 1.5\times10^{-1}, 1\times10^{-1}\}$. We find that the best $\tau$ for ResNet-110 (w/ BN), DenseNet-100 (w/ BN), and WRN-28-10 (w/ BN) are $1\times10^{-1}, 1.5\times10^{-1},$ and $2.5\times10^{-1}$, respectively. For architectures without BN layers, we try $\tau$ from smaller values  $\{3\times10^{-2}, 2.5\times10^{-2}, 2\times10^{-2}, 1.5\times10^{-2}, 1\times10^{-2}\}$. The best $\tau$ for ResNet-110 (w/o BN), DenseNet-100 (w/o BN), and WRN-28-10 (w/o BN) are $1.5\times10^{-2}, 1.5\times10^{-2},$ and $3\times10^{-1}$, respectively. To speed up NIO for a fast initialization, we use the batch setting for all architectures with and without BN layers. We set the number of sub-batches $D$ as 2 under a batch size of $B=128$. To relieve the violation of the sample-wise analysis, we adopt an overlap between the two sub-batches. Specifically, the best overlap ratios $r$ for ResNet-110 (w/ BN), DenseNet-100 (w/ BN), and WRN-28-10 (w/ BN) are $0.6, 0.6,$ and $0.7$, respectively. The best overlap ratios $r$ for ResNet-110 (w/o BN), DenseNet-100 (w/o BN), and WRN-28-10 (w/o BN) are $0.7, 0.75,$ and $0.8$ respectively. Finally, for the upper bound of the gradient norm $\gamma$, we follow the choices suggested in \cite{zhu2021gradinit}. The best $\gamma$ for ResNet-110 (w/ BN), ResNet-110 (w/o BN), DenseNet-100 (w/ BN), DenseNet-100 (w/o BN), WRN-28-10 (w/ BN), and WRN-28-10 (w/o BN) are $5, 4, 3, 3.5, 2,$ and $2$, respectively for CIFAR-10.
CIFAR-100 has $10^2$ classes, so the initial cross entropy loss will be about 2 times as large as the one on CIFAR-10 for the same architecture. Following the principle in \cite{zhu2021gradinit}, we accordingly set $\gamma$ 2 times as large as that for CIFAR-10. NIO is iterated until all the training samples are covered.

\paragraph{Training details.} 
After NIO, we train these networks on CIFAR-10 and CIFAR-100 for 500 epochs with a batch size of 128 on a single GPU. 
The initial learning rate is 0.1 and decays after each iteration following the cosine annealing learning rate schedule without restart  \cite{loshchilov2016sgdr}.
We set weight decay to $10^{-4}$ in all experiments.
For networks without BN layers, we apply gradient clipping (with a maximum gradient norm of 1) to prevent from gradient explosion. 
Following the standard data augmentation schemes, we adopt random cropping, random flipping, 
and cutout \cite{devries2017improved} for all networks. We do not use dropout in any of our experiments. 

\subsection{On ImageNet}
\label{ap23}
\paragraph{Architecture Details.}
To show the extensibility of NIO to large datasets and novel architectures, we also conduct experiments on ImageNet with ResNet-50 and Swin-Transformer. Specifically, we use the standard ResNet-50 architecture that is stacked by 3-layer bottleneck blocks. BN layers are enabled as default. For experiments of Swin-Transformer, we adopt the Swin-Tiny configuration, where the window size $M$ is 7, the query dimension of each head $d$ is 32, and the expansion ratio of each MLP block is 4. The initial channel number is 96, and the numbers of blocks for four stages are $\{ 2,2,6,2\}$. We follow the official implementation in \cite{liu2021swin}.

\paragraph{NIO Configurations.}
Since the networks on ImageNet have more parameters than the architectures on CIFAR10/100, we adopt a relatively smaller learning rate $\tau$. Concretely, we try $\tau$ from $\{5\times10^{-2} ,1\times10^{-2},5\times10^{-3}, 3\times10^{-3}, 1\times10^{-3}\}$ and choose $\tau = 3\times10^{-3}$ for both ResNet-50 and Swin-Transformer. 
We set the number of sub-batches $D= 4$ under the batch size $B=64$ and use an overlap ratio of $r=0.2$ for both architectures. Following the previous principle, we change the upper bound of the gradient norm $\gamma$ three times as large as that for CIFAR-10 due to the increased number of categories. Finally, we run NIO for 100 iterations on ImageNet to keep the initialization process efficient.



\paragraph{Training Details.} 
Following the standard implementations, we train ResNet-50 for 100 epochs and Swin-Transformer for 300 epochs. The batchsize is 256 for ResNet-50 and 1024 for Swin-Transformer among 8 NVIDIA A-100 GPUs. 
ResNet-50 is trained by the SGD optimizer with a weight decay of $10^{-4}$ and a momentum of 0.9. Swin-Transformer is trained by the AdamW optimizer with a weight decay of 0.05 and a momentum of (0.9, 0.999).
The initial learning for ResNet-50 is 0.1, and is divided by a factor of 10 at epoch 30, 60, and 90. The base learning rate for Swin-Transformer is $1\times10^{-3}$, and follows the cosine annealing learning rate schedule without restart. For Swin-Transformer with warmup enabled, a linear learning rate warmup is used for the first 20 epochs of training. Swin-Transformer adopts gradient clipping with a maximum gradient norm of 5, and the standard augmentation schemes, such as Mixup \cite{zhang2017mixup}.


\section{More Explanations and Discussions}
\label{explanation}

As suggested by the reviewers of this paper, we offer more explanations and discussions here for some contents in the main paper.

1. Explanations for Figure \ref{fig1}.

In Figure \ref{fig1} (c) and (d), it seems that an initialization with a smaller cosine similarity (larger angle) is closer to the global optimum of the two samples. But we should note that:

(1) The optimum of a given network is not unique. It is dependent on its initialization. The model with different initialization points will converge to different optimal parameters, even though their performances may be close. So, what we do here is to look for an initialization whose sample-wise optimization paths are more consistent (smaller angle), instead of the inverse way---looking for a point closer to the global optimum of some given sample-wise optima and landscapes, which is impossible for a real network. The larger optimization path consistency induces a smaller gradient variance, whose benefits have been supported by prior studies \cite{liu2019variance,zhuang2020adabelief}. 

(2) The illustration in Figure \ref{fig1} is a toy example with only two samples and simple landscapes. In a real network, the landscapes are highly non-convex in a high dimension, and there are a larger number of training samples. In this case, the global optimum does not necessarily lie in the convex hull. The point with a larger angle to sample-wise optima is not ensured to be closer to the global optimum.

In conclusion, an initialization with more consistent sample-wise optimization paths is more favorable. Our aim is to look for such initialization, instead of the global optimum of given sample-wise landscapes. Figure \ref{fig1} (c) and (d) are toy examples that are only used to show that the metric Eq. (\ref{gradsign}) is agnostic of initialization, while our Eq. (\ref{eq2}) reflects the optimization path consistency and is a function of initialization.

2. Explanations for some concepts.

Sample-wise optimization refers to training the model only on each single sample. So, the optimization path for sample $i$ can be characterized by $\theta_i^*-\theta_0$, where $\theta_0$ is the initialized point, and $\theta_i^*$ is the converged optimum of only training on sample $i$. Because all sample-wise optimization paths have the same starting point $\theta_0$, we use the averaged Cosine similarity to measure their consistency. The consistency of sample-wise optimization path can be formulated as $\frac{1}{n^2}\sum_{i,j}\cos\angle\left(\theta_i^*-\theta_0, \theta_j^*-\theta_0\right)$. So, if the angle between the paths from the initialization $\theta_0$ to each local optimum $\theta_i^*$ is small, we will have more consistent sample-wise optimization paths. It is approximated by our GradCosine in Line 6 in Algorithm \ref{alg1}. 

It is shown that our proposed Eq. (\ref{eq2}) reflects the optimization path consistency and is a function of initialization $\theta_0$. The aim of our NIO is to look for an initialization $\theta_0$ that maximizes GradCosine.

3. Discussions about the rationality of the approximation in Eq. (\ref{eq4}).

Note that $\theta_i^*$ in Eq. (\ref{eq4}) is not the global optimum. It is the local optimum for sample $i$. If we optimize a model on only one training sample, it is very easy to finish the training. Only several iterations are needed to attain a zero loss. So, we make the first-order approximation for the sample-wise optimization, \emph{i.e.}, the sample-wise optimum can be reached via only one step of gradient descent.

4. Discussions about the comparison with GradSign \cite{zhang2021gradsign}.

In GradSign \cite{zhang2021gradsign}, the metric of sample-wise optima density (Eq. (\ref{gradsign}) in our paper) is proposed and proved to be an upper bound of training and generalization error. Because sample-wise optima are not tractable, they approximate it by counting the gradient sign, which is non-differentiable. The metric is used to rank neural architectures without training for a neural architecture search purpose.

However, the goal of our work is to develop a method for initialization optimization, instead of architecture search. So, our theoretical metric needs to be differentiable and be a function of initialization. The differences between \cite{zhang2021gradsign} and our work are summarized as follows:

(1) The metric Eq. (\ref{gradsign}) proposed in \cite{zhang2021gradsign} is agnostic of initialization (as shown in Figure \ref{fig1} (c) and (d)). Its approximated quantity, GradSign, is non-differentiable. So, it cannot be used for initialization optimization. In contrast, our metric Eq. (\ref{eq2}) reflects the optimization path consistency as a function of initialization. It is approximated by the differentiable GradCosine, and thus can serve for our initialization optimization purpose.

(2) Our method considers optimization path consistency, which is intimately related to a favorable training dynamic. The larger optimization path consistency induces a smaller gradient variance, whose benefits have been supported by prior studies \cite{liu2019variance,zhuang2020adabelief}. See more details in the ``Relations to Favorable Training Dynamic'' paragraph. As a comparison, \cite{zhang2021gradsign} does not enjoy this advantage.


(3) Although both metrics Eq. (\ref{gradsign}) and Eq. (\ref{eq2}) are related to model performance, because our work and \cite{zhang2021gradsign} aim for different tasks, the other parts of our work including the GradCosine quantity, the objective of our neural initialization optimization, and the algorithm to solve it, are original and have no similarity with \cite{zhang2021gradsign}.

\end{document}